\documentclass[runningheads]{llncs}
\usepackage{cite} 
\usepackage{graphicx,wrapfig}
\usepackage{subcaption}
\usepackage{amsmath}
\usepackage{bm}
\usepackage{footnote}
\usepackage{hyperref}
\begin{document}
\title{Multi-frame Super-resolution from Noisy Data}
%
%
\author{Kireeti Bodduna \and
Joachim Weickert \and
Marcelo C{\'a}rdenas}
\authorrunning{Bodduna et al.}

%
\institute{Mathematical Image Analysis Group, 
Faculty of Mathematics and Computer Science, 
Saarland University, 66041 Saarbr{\"u}cken, Germany.
\email{\{bodduna,weickert,cardenas\}@mia.uni-saarland.de}}
\maketitle              
\begin{abstract}
Obtaining high resolution images from low resolution data with
clipped noise is algorithmically challenging due to the ill-posed 
nature of the problem. So far such problems have hardly been tackled, 
and the few existing approaches use simplistic regularisers. We 
show the usefulness of two adaptive regularisers based on 
anisotropic diffusion ideas: Apart from 
evaluating the classical edge-enhancing anisotropic diffusion 
regulariser, we introduce a novel non-local one with one-sided 
differences and superior performance. It is termed sector diffusion. 
We combine it with all six variants of the classical super-resolution 
observational model that arise from permutations of its three
operators for warping, blurring, and downsampling.
Surprisingly, the evaluation in a practically relevant noisy 
scenario produces a different ranking than the one in the 
noise-free setting in our previous work (SSVM 2017).

\end{abstract}
%
%
\section{Introduction}
\label{sec:1}
Super-resolution (SR) is an image processing technique designed 
to overcome the resolution limits of cameras. Generating a high 
resolution (HR) image from one single low resolution (LR) image is
referred to as single-frame super-resolution \cite{GSI09,MO2008,YWHM10}. 
In this work, we concentrate on multi-frame super-resolution, 
where information from multiple LR images is fused into a single HR 
image \cite{PVS2006,MO2008,YZS2012,FREM2004a,LHR2015,TLAN2002,
KDV2012,MKDD2015,LFCS2005,LMZBBGAC2013,
DDICGDPSI2015,KBPS2011,TLAN2001}.

In bio-medical and bio-physical applications we 
encounter images that possess a significant amount of noise.
Multi-frame super-resolution in the presence of noise is thus 
practically relevant and also a very challenging research field. 
Algorithms that are designed to solve this problem 
compute derivative information on noisy data which showcases 
the ill-posed nature of the problem. In view of these algorithmic 
challenges, it is not surprising that very little efforts have been 
put into obtaining high resolution images from noisy low resolution
data. 

Deep learning-based methods are less suitable 
for bio-physical applications like electron microscopic imaging 
due to three main reasons: Firstly, there is very little ground truth 
data available. Secondly, the raw noise type is not well understood, 
unlike normal cameras. Finally, this imaging pipeline employs a huge 
amount of steps to obtain the final structure of the specimen under 
observation. After each one of these steps, the noise type changes. 
This makes deep-learning models that are specifically trained 
for a particular kind of noise, sub-optimal.

In the present paper 
we specifically aim at reconstructing a HR image from its
LR versions that have been corrupted by additive white 
Gaussian noise (AWGN).

\subsubsection{Formalisation of the Problem.}
For multi-frame super-resolution, we want to find a HR image $\bm{u}$ 
of resolution $N_H = H_1 \times H_2$ from $N$ low resolution images 
$ {\left\{\bm{f}_{i}\right\}} _{i=1}^{N}$ of 
resolution $N_L = L_1 \times L_2$. 
The low resolution
images are assumed to be degraded versions of the real world HR scene. 
The standard formulation of the relation
between the high resolution scene and its 
LR realisations is \cite{EF1997}
\begin{equation}
\label{eq:1}
\bm{f}_{i} = \bm{D}\bm{B}\bm{W}_i\bm{u} + \bm{e}_i.
\end{equation} 
In this observational model, 
we express the motion of the objects in the image  
using the warping operator $\bm{W}_i$ (size: $N_H \times N_H$). 
The operator $\bm{B}$ (size: $N_H \times N_H$) denotes the 
blur due to the point spread function of the camera. 
We represent the downsampling of the HR scene by the 
camera detector system using $\bm{D}$ (size: $N_L \times N_H$).
The vector $\bm{e}_i$ depicts the noise (error) acquired due to 
the imaging system. The operators $\bm{B}$ and $\bm{D}$
do not have an index $i$ as we assume the same camera conditions
for all images.

The standard model (\ref{eq:1}), however, has a disadvantage: 
The operator $\bm{W}_i$ acts on the high-resolution 
scene $\bm{u}$. Hence, the model assumes that we have 
motion information at the high-resolution scale. In practice, 
we just have the downsampled and blurred images $\bm{f}_{i}$ 
at our disposal.  
Motion at high-resolution must be approximated by 
upsampling the one computed on a lower resolution. 
Thus, the following question arises: 
Can one improve the practical performance of the SR approach
by permuting the order of the operators? The seminal work of 
Wang and Qi \cite{WQ2004} and the paper by Bodduna and 
Weickert \cite{BW2017} made progress in this direction. 
In \cite{BW2017}, the authors tried to evaluate the SR observational 
model in a noise-free scenario by studying the six different 
alternatives that arise from permutations of its three operators 
$\bm{D}$, $\bm{B}$ and $\bm{W}_i$. This has led 
to improvements in terms of both quality and speed, the former 
of which was  also observed in \cite{WQ2004}. However, such 
an evaluation is missing for the practically relevant scenario 
with noisy images. Our paper will address this problem.

Moreover, there is a second problem: For super-resolution of noisy
data, the ideal observational model in \eqref{eq:1} should be 
stabilised with a regulariser. In most cases, this is
done by embedding it into the following quadratic energy minimisation 
framework:
\begin{equation}
\label{eq:2}
E(\bm{u}) = \frac{1}{2} \sum_{i=1}^{N} | \bm{DBW}_i\bm{u}
-  \bm{f}_i |^2 + \frac{1}{2} \alpha  | \bm{A} \bm{u}|^2.
\end{equation}
Here, $\bm{A}$ is a discrete approximation of the continuous 
gradient operator, $\alpha$ is the regularisation constant, and 
$|\cdot|$ denotes the Euclidean norm.
The first term is the data term that encapsulates the observational 
model. 
The second one serves as smoothness term which eliminates noise. 
Minimising (\ref{eq:2}) by setting its gradient to zero gives
   \begin{equation}
   \label{eq:3}
    \sum_{i=1}^{N} \bm{W}_i^\top \bm{B}^\top \bm{D}^\top 
    (  \bm{DBW}_i \bm{u} -  \bm{f}_i ) - \alpha 
    \bm{A}_{\rm{HD}} \bm{u} = \bm{0},
   \end{equation} 
where $\bm{A}_{\rm{HD}} = \bm{A}^\top\bm{A}$ is the discrete 
approximation of the continuous Laplacian operator. 
In this paper, we use a Gaussian blur kernel, such that 
$\bm{B}^\top$ equals $\bm{B}$. We denote the upsampling and downsampling 
matrices by $\bm{D}^\top$ and $\bm{D}$, respectively. 
The operator $\bm{W}_i$ represents 
forward warping, while $\bm{W}_i^\top$ encodes 
backward registration. 
The explicit gradient descent scheme with parameters 
$\tau$ (the time step size) and $k_{\textrm{max}}$ (the 
number of iterations) to solve Equation (\ref{eq:3}) is given 
by
\begin{equation}
\label{eq:4}
\begin{split}
\bm{u}^{k+1} & = \bm{u}^k + 
\tau \Big( \alpha \bm{A}_{\rm{HD}} \bm{u}^{k} 
 - \sum_{i=1}^{N} \bm{W}_i^\top \bm{B}^\top \bm{D}^\top 
(  \bm{DBW}_i \bm{u}^k -  \bm{f}_i ) \Big) .
\end{split}
\end{equation}
In this evolution equation, $\bm{A}_{\rm{HD}}$ acts as the denoiser.
However, such a noise elimination scheme uses a simple homogeneous 
diffusion process that also blurs important structures. 
As far as the usage of diffusion-based regularisers for super-resolution 
is concerned, only a few papers with simplistic models are available. 
Thus, it is highly desirable to introduce more advanced structure
preserving regularisers. This is our second challenge.

\subsubsection{Our Contribution.} 
To address the first problem, we investigate the performance of 
all  permutations of the standard observational model 
in an AWGN setting that is clipped to the dynamic range $[0,255]$. 
This practically relevant noise model covers over- and under-exposed 
image acquisition conditions.

To incorporate structure-preserving regularisers, we start with
replacing the homogeneous diffusion operator  
by the classical model of edge-enhancing aniso\-tropic diffusion (EED)
\cite{We94e}. Although this model is around since a long time,
its performance for super-resolution has not been examined so far.
Moreover, we also introduce a more sophisticated non-local 
anisotropic model that offers better structure preservation and 
superior noise elimination than EED.
We call it sector diffusion (SD). It differs from all other diffusion 
models by the fact that it is fully based on one-sided derivatives. 

We first compare the denoising performance of EED and SD
for real-world images with clipped-AWGN, before we embed them
as regularisers in the SR framework.   
We deliberately do not evaluate popular denoising methods such as 
3D block matching \cite{DFKE07} and non-local Bayes \cite{LBM2013}:  
Most of these techniques rely heavily on a correct noise model,
which renders them inferior for clipped noise, in particular with
large amplitudes.

\subsubsection{Paper Structure.} Our paper is organised as 
follows: We introduce our novel sector diffusion model in 
Section~\ref{sec:modeling}. Here, we also review various 
super-resolution observational 
models and the EED-based image evolution equation.
In Section~\ref{sec:exp}, we present several denoising and 
SR reconstruction experiments along with some discussions.
Finally, in Section~\ref{sec:conclusion} we conclude with a summary 
about robust multi-frame SR reconstruction as well as 
an outlook on future work.
\section{Modeling and Theory}
\label{sec:modeling}
In this section, we first review the various 
possible permutations of the super-resolution observational model 
in \eqref{eq:1}. Afterwards, we introduce the different regularisation 
schemes utilised for both denoising and SR reconstruction purposes.
\vspace{-1em}
\begin{table}[t]
\begin{center}
\caption{\small \textbf{Left:} The seven SR observational models.
\textbf{Right:} Parameter settings for optical flow calculation.
We have two model parameters: $\alpha_{OF}$ (smoothness parameter) 
and $\sigma_{OF}$ (Gaussian pre-smoothing). 
Numerical parameters are chosen as $\eta = 0.95$ (downsampling factor), 
$\eta_1 = 10$ (inner fixed point iterations), $\eta_2 = 10$ 
(outer fixed point iterations) and $\omega = 1.95$ 
(successive over-relaxation parameter).}
\label{tab:SR_models}
\vspace{0.5em}
\begin{minipage}{0.45\textwidth}
\footnotesize
\flushright
\begin{tabular}{l r}
\hline \newline
Model & \;\;\; Equation \\
\hline
M1   & $\bm{DBW}_i\bm{u} + \bm{e}_i= \bm{f}_i$  \\ [2.5pt]
M2   & $\bm{DW}_i\bm{B}\bm{u} + \bm{e}_i= \bm{f}_i$  \\ [2.5pt]
M3   & $\bm{BDW}_i\bm{u} + \bm{e}_i= \bm{f}_i$  \\ [2.5pt]
M4   & $\bm{W}_i\bm{DB}\bm{u} + \bm{e}_i= \bm{f}_i$  \\ [2.5pt]
M5   & $\bm{BW}_i\bm{D}\bm{u} + \bm{e}_i= \bm{f}_i$ \\ [2.5pt]
M6   & $\bm{W}_i\bm{BD}\bm{u} + \bm{e}_i= \bm{f}_i$ \\ [2.5pt]
M2.1 \; & $\bm{B}\bm{u} + \bm{e}_i = 
\bm{D^\top W^\top}_i\bm{f}_i$ \\
\hline
\end{tabular}
\end{minipage}
\hspace{1em}
\begin{minipage}{0.49\textwidth}
\vspace{-1.4em}
\footnotesize
\begin{tabular}{lrr}
\hline 
\multicolumn{1}{p{1.5cm}}{Dataset}
& \multicolumn{1}{p{0.5cm}}{$\sigma_{OF}$}
& \multicolumn{1}{p{0.5cm}}{$\;\;\; \alpha_{OF}$}\\\hline
\noindent Text1 & 2.6 & \;\; 13.3  \\ [2pt]
Text2 & 1.0 & 15.6  \\ [2pt]
Text3 & 2.3 & 6.3   \\ [2pt]
House1 & 3.8 & 13.5 \\ [2pt]
House2 & 1.2 & 17.0 \\ [2pt]
House3 & 2.7 & 16.5 \\ 
\hline
\end{tabular}
\end{minipage}
\end{center}
\vspace{-5pt}
\end{table}
\subsection{Super-resolution Observational Models}
Table \ref{tab:SR_models} shows the various permutations of the 
original observational model M1 \cite{BW2017}. 
While models M2-M6 depict 
the five other possible permutations, M2.1 represents a technique 
that is derived from M2. The motivation behind the modelling 
of M1-M6 is quality reasons. M2.1, on the other hand, 
is designed to exploit the precomputable nature of the 
term on the right hand side of the corresponding equation. 
Such a design is faster than any of the other models.
\subsection{Edge-enhancing Diffusion}
\label{sec:EED}
Edge-enhancing diffusion was proposed by Weickert \cite{We94e} with 
the goal to enhance smoothing along edges while inhibiting it 
across them. To achieve this, one designs a diffusion tensor $\bm{D}$
with eigenvectors $\bm{v}_1$ and $\bm{v}_2$ that are
parallel and perpendicular to a Gaussian smoothed image gradient. 
This is followed by setting the eigenvalue corresponding to the 
eigenvector perpendicular to the gradient to one, indicating 
full flow. The eigenvalue corresponding to the eigenvector  
parallel to the gradient is determined by a  
diffusivity function. Using this idea, one can inhibit 
smoothing across edges. The following is the continuous
mathematical formulation of the evolution of image $u$ under 
EED:
\begin{equation}
\label{eq:EED}
\partial_{t} u = \textrm{div}(\bm{D}(\bm{\nabla} u_{\sigma}) 
\bm{\nabla} u) ,
\end{equation}
\begin{equation}
\bm{D}(\bm{\nabla} u_{\sigma}) = g(|\bm{\nabla} u_{\sigma}|^2) 
\cdot \bm{v}_1\bm{v}_1^{T} 
+ \ 1 \cdot \bm{v}_2\bm{v}_2^{T},
\end{equation}
\begin{equation}
\bm{v}_1 \parallel \bm{\nabla} u_{\sigma} , \ \
|\bm{v}_1| = 1 \ \ \ \textrm{and}  \ \ \ \bm{v}_2
\perp \bm{\nabla} u_{\sigma} , \ \ |\bm{v}_2| = 1.
\end{equation}
Here, div is the 2D divergence operator and $\bm{\nabla} u$ 
the spatial gradient. The Gaussian-smoothed image 
is $u_\sigma$. Computing the gradient on $u_{\sigma}$  
makes the diffusion process robust under the presence of noise. 
Both EED and SD evolution equations are initialised 
with the noisy image $\bm{f}$.
Finally, the diffusivity function $g(x)$ is chosen as \cite{We97}
\begin{equation}
\label{diffusivity}
g\left( x \right) = 1 - \text{exp}\left( 
\frac{-3.31488}{\left(\frac{x}{\lambda}\right)^8} \right).
\end{equation} 
Thus, by replacing the Laplacian $\bm{A}_{\rm{HD}}$ 
in \eqref{eq:4} with the space discrete 
version $\bm{A_{\textrm{EED}}}$ of the EED operator in 
\eqref{eq:EED}, 
we arrive at the EED-based scheme 
for reconstructing the high resolution scene:
\begin{equation}
\label{eq:EED_SR}
\begin{split}
\bm{u}^{k+1} & = \bm{u}^k + 
\tau \Big( \alpha (\bm{A_{\textrm{EED}}}(\bm{u}^k))  
 - \sum_{i=1}^{N} \bm{W}_i^\top \bm{B}^\top \bm{D}^\top 
(  \bm{DBW}_i \bm{u}^k -  \bm{f}_L^i ) \Big) .
\end{split}
\end{equation}
The details regarding the discretisation of the EED operator 
can be found in \cite{WWW2013}.
\subsection{Sector Diffusion}
\label{sec:SD}
\subsubsection{Continuous Model.}
Our goal is to design a diffusion method with a higher adaptation 
to image structures than previous anisotropic models such as EED.
To this end, we start with Weickert's integration model from \cite{We94a}: 
\begin{equation}\label{ani diff}
\partial_t u(\bm{x},t) = \frac{1}{\pi}\int_0^{\pi}\partial_\theta 
\left( g\left(\partial_\theta u_\sigma\right)\partial_\theta u \right) 
d\theta.
\end{equation}
Here, $\partial_\theta$ stands for the directional derivative in 
the direction represented by angle $\theta$, $g$ is the diffusivity 
function and $u_\sigma$ denotes a convolution of $u$ with a 
Gaussian of standard deviation $\sigma.$
This model considers each orientation separately and is thus capable of 
diffusing along edges, but not across them. 

In order to improve its structure adaptation even further, we replace
the directional derivatives by one-sided directional derivatives and
integrate over $[0,2\pi]$ instead of $[0,\pi]$: 
\begin{equation}
\label{new_anisotropic_diffusion}
\partial_t u(\bm{x},t) = \frac{1}{2\pi}\int_0^{2\pi}\partial^+_\theta 
\left( g \left(\partial^+_\theta u_\sigma^{\theta}\right)
\partial^+_\theta u \right) 
d\theta.
\end{equation}
Here, $u_\sigma^{\theta}$ represents a one-sided smoothing of $u$ in the 
orientation given by the angle $\theta,$ and $\partial^+_\theta$ denotes 
a one-sided derivative in the same orientation. In contrast to the 
usual Gaussian smoothing applied in (\ref{ani diff}), this one-sided 
smoothing allows the filter to distinguish two different derivatives for 
a given direction: One in the orientation of $\theta,$ and the other in 
the orientation of $\theta + \pi.$ 
A formal definition of these concepts can be realised by considering the 
restriction of $u$ to the corresponding ray starting at 
$\bm{x},$ in the orientation of each $\theta.$ Namely, 
for fixed $\bm{x},t,\theta,$ we consider $u(h;\bm{x},t)
:=u(\bm{x} + h(\cos(\theta),\sin(\theta))^{T},t),$ for $h\in[0,\infty]$. 
Then, the one-sided directional derivative $\partial^{+}_\theta$ is 
formally defined as 
\begin{equation}\label{eqCorner-cont-nonlocal}
\partial^{+}_\theta u := \lim_{h\rightarrow 0^{+}}\frac{u(h;\bm{x},t) - 
u(\bm{x},t)}{h}.
\end{equation}
To our knowledge, diffusion filters that are explicitly based on
one-sided directional derivatives have not been described in the 
literature so far. 

In order to introduce a second alteration of model (\ref{ani diff}), 
we incorporate the concept of non-locality. This leads to 
\begin{gather}
\begin{align*}
\footnotesize
\centering
\begin{split}
\hspace{-5em}
\partial_{t} {u(\bm{x},t)} = \! 
\int\limits_{B_{x,\rho}}  \! \! \! J(|\bm{y}-\bm{x}|) 
 \; g \left(\frac{{u}_\sigma(\bm{y};\bm{y}-\bm{x}) - {u}_\sigma(\bm{x};
\bm{y}-\bm{x})}{|\bm{y} - 
\bm{x}|}\right) 
 \left(u(\bm{y}) - u(\bm{x})\right)\,d\bm{y}.
\end{split}
\end{align*}
\end{gather}
Here, $B_{x,\rho}$ denotes the disc with center $x$ and radius $\rho.$
The diffusivity \textit{g} has already been defined in 
\eqref{diffusivity}. Also, the function $J(s)$ is a slightly 
Gaussian-smoothed version of $F(s):=\frac{1}{s^2}$. Moreover, we
assume that it decreases fast but smoothly to zero 
such that $J(s)=0$ for $|s|\geq\rho.$ 
The slight Gaussian smoothing of $F$ is required 
for avoiding the singularity of $J$ as $s\rightarrow 0.$ 
The value ${u}_\sigma(\bm{z};\bm{y}-\bm{x})$ corresponds 
to a one-dimensional Gaussian smoothing of $u$ 
inside the segment $\lambda_{\bm{x}\bm{y}}(s):=
\left\{\bm{x} + s\frac{(\bm{y} - \bm{x})}{|\bm{y} - \bm{x}|}:\, 
s\in[0,\rho]\right\}$ evaluated at $\bm{z}.$
This idea of making the diffusivity dependent on values 
inside an orientation dependent segment 
determines the structure preservation capabilities of the 
model. In the next paragraph we will see how to translate this 
non-local filter into a space-discrete version by dividing the 
disc $B_{x,\rho}$ into sectors. This explains the name {\em sector 
diffusion.}
\begin{table}[t] 
\captionof{table}{\small 
 MSE values of denoised images 
 including parameters used.  L40 stands for Lena image with 
 $\sigma_{\textrm{noise}}$ = 40.  B, H, P denote Bridge, House and 
 Peppers respectively.}
\vspace{0.5em}
\setlength{\tabcolsep}{3pt}
\footnotesize
\centering
\hspace{5mm} EED \hspace{30mm} SD \hspace{5mm} \\[5pt]
\begin{tabular}{ l r r r r }
 \hline 
 Image & $\sigma$ & $\lambda$ 
 & $k_{\textrm{max}}$ & MSE \\
 \hline 
 L40 & 1.2 & 7.5 & 34 & 98.67  \\  
 L60 & 1.8 & 5.0 & 63 & 156.24 \\  
  \vspace{0.5em}
 L80 & 2.0 & 4.6 & 87 & \;\;230.28  \\  
 B40 & 0.9 & 14.4 & 12 & 294.32    \\  
 B60 & 1.1 & 13.4 & 20 & 418.71  \\  
               \vspace{0.5em}
 B80 & 1.4 & 10.4 & 28 & 514.23 \\   
 H40 & 0.9 & 11.1 & 34 & \textbf{96.62} \\ 
 H60 & 1.1 & 12.1 & 33 & 167.72 \\ 
  \vspace{0.5em} 
 H80 & 1.8 & 5.8 & 72 & 247.09 \\    
 P40 & 1.2 & 8.1 & 28 & 102.97 \\ 
 P60 & 1.7 & 5.6  & 51 & 200.31 \\
 P80 & 1.9 & 5.1 & 68 & 353.61 \\    
 \hline 
 \end{tabular}
 \hspace{4mm}
 \begin{tabular}{ r r r r}
 \hline 
 $\sigma$ & $\lambda$ & $k_{\textrm{max}}$ & \;\;\;\;\; MSE \\
 \hline 
 0.6 & 3.1 & 7 & \textbf{92.99} \\  
 0.6 & 3.3 & 11 & \textbf{138.48}\\  
  \vspace{0.5em}
 0.6 & 2.9 & 18 & \textbf{180.66} \\  
 0.5 & 3.3 & 4 & \textbf{261.62}  \\  
 0.5 & 4.1 & 6 & \textbf{360.87} \\  
               \vspace{0.5em}
 0.6 & 4.0 & 9 & \textbf{436.60} \\   
 0.7 & 2.6 & 9 & 104.31\\ 
 0.7 & 2.7 & 14 & \textbf{152.24} \\ 
  \vspace{0.5em} 
 0.6 & 2.7 & 19 & \textbf{207.65} \\    
 0.6 & 2.1 & 10 & \textbf{86.57} \\ 
 0.6 & 1.8 & 19 & \textbf{133.19}\\
 0.6 & 1.7 & 30 & \textbf{188.86} \\    
 \hline 
 \end{tabular}
 \label{tab:denoising}
\end{table}

\subsubsection{Discrete Model.}
In order to properly adapt our filter to the local image structure, 
we first divide a disc shaped neighborhood $B_{i,\rho}$
of radius $\rho$ centered around pixel $i$, into $M$ sectors. 
With the objective of reducing interactions between regions of 
dissimilar grey values we employ robust smoothing 
within these sectors. This mirrors the continuous modelling idea 
of smoothing within the segments $\lambda_{\bm{x}\bm{y}}$.   
The final design objective is that we employ one-sided  
finite differences instead of central differences for discretisation 
purposes. The latter have a property of smoothing over the 
central pixel, thus destroying image structures. This idea is 
again a direct consequence of considering orientations rather 
that directions, in the continuous model. 
With these motivations in mind, we define the space-discrete 
formulation of the sector diffusion model as
\begin{align}\label{sec_evol}
\frac{du_i}{dt} = 
\bm{A_{\textrm{SD}}}(u)
=
\sum_{\ell=1}^{M} \sum_{\substack{j \in S_\ell} } 
g_{i,j} \cdot
\frac{{u}_j - {u}_i}{|{\bm{x}}_{j} - {\bm{x}}_{i}|^2}.
\end{align}
Here, $g_{i,j} = g \left(\frac{{{u}_\sigma}_{j\ell} - 
{{u}_\sigma}_{i\ell}}{|{\bm{x}}_{j} - {\bm{x}}_{i}|}\right)$,
$S_\ell$ is the set of pixels within a particular 
sector $\ell$, and ${\bm{x}}_i$ and ${\bm{x}}_j$ denote the position 
of the pixels $i$ and $j$ in the image grid. 
The sector-restricted smoothing is defined as
\begin{equation}
\label{neigh_smoothing}
{{u}_\sigma}_{j\ell} = \frac{1}{c}\sum_{\substack{k \in S_\ell }} 
h(k,j,\sigma)u_k.
\end{equation}
Here, $c$ is a normalisation constant and 
\begin{equation}
\label{a_func}
h(k,j,\sigma)= 
\text{exp}\left(\frac{-|{\bm{x}}_k - {\bm{x}}_j|^2}{2\sigma^2}\right).
\end{equation} 
Similar to EED, we can now define the SD-based SR framework as
\begin{equation}
\begin{split}
\bm{u}^{k+1} & = \bm{u}^k + 
\tau \Big( \alpha (\bm{A_{\textrm{SD}}}(\bm{u}^k))  
 - \sum_{i=1}^{N} \bm{W}_i^\top \bm{B}^\top \bm{D}^\top 
(  \bm{DBW}_i \bm{u}^k -  \bm{f}_L^i ) \Big) .
\end{split}
\label{eq:SD_SR}
\end{equation}
\section{Experiments and Discussion}
\label{sec:exp}

In this section, we first evaluate EED and SD in terms of their 
denoising as well as SR regularisation capability. Then we choose 
the best of the two as regulariser for evaluating the operator orders 
in the SR observational model. 
\begin{table*}[t]
 \captionof{table}{\small
 MSE values of SR reconstructed images 
 including parameters used.  T2 stands for Text2 dataset with 
 ground truth optical flow, while T2-S was computed using 
 sub-optimal calculated flow. Ground truth image size 
 for Text: 512 $\times$ 512. T1-T3 represent images downsized by 
 factors 1, 2 and 3, respectively. Image size for 
 House: 256 $\times$ 256. H1-H3 represent images downsized by 
 factors 1, 1.5 and 2, respectively. Every dataset has 30 images 
 each, with the last of them being the reference frame for 
 registration.}
 \vspace{5pt}
\setlength{\tabcolsep}{3pt}
\footnotesize
\centering
\hspace{5mm} EED \hspace{50mm} SD \hspace{5mm} \\[5pt]
\begin{tabular}{ l r  r r  r r r}
 \hline 
 Dataset & $\sigma$ & $\;\;\sigma_B$ & $\;\lambda$ 
 & $\alpha$ & $k_{\textrm{max}}$ & \;\;\;MSE \\
 \hline 
 H1 & 0.6 & 0.8 & 11.0 & 118.0 & 37 & 110.45  \\  
 H2 & 0.7 & 0.5 & 12.0 & 120.0 & 9 & 162.64   \\  
 H2-S & 0.7 & 0.5 & 13.0 & 115.0 & 9 & 172.94 \\  
\vspace{0.5em}H3 & 0.6 & 0.4 & 14.0 & 127.0 & 48 & 201.91  \\  
 T1   & 1.0 & 1.1 & 9.0 & 14.0 & 136& 164.72 \\  
 T2   & 1.3 & 0.9 & 7.0 & 18.0 & 11 & 397.09 \\  
 T2-S & 1.3 & 1.0 & 7.0 & 18.0 & 14 & 510.80 \\  
 T3   & 1.2 & 0.4 & 7.0 & 14.0 & 13 & 674.65 \\   
 \hline 
 \end{tabular}
 \hspace{2mm}
\begin{tabular}{ r r  r r  r r r}
 \hline 
 $\sigma$ & $\;\;\sigma_B$ & $\;\;\;\;\lambda$ 
 & $\;\;\;\;\alpha$ & $k_{\textrm{max}}$ & \;\;\;\;\;MSE \\
 \hline 
 0.9 & 1.0 & 2.0 & 1.6 & 17 & \textbf{83.12}  \\  
 0.8 & 0.7 & 1.7 & 5.3 & 17 & \textbf{133.35} \\  
 0.9 & 0.8 & 1.8 & 4.5 & 17 & \textbf{141.62} \\  
\vspace{0.5em}0.6 & 0.8 & 2.3 & 2.9 & 49 & \textbf{161.26}  \\  
 0.6 & 1.1 & 3.0 & 0.3 & 48 & \textbf{158.50} \\  
 0.6 & 1.0 & 2.7 & 0.6 & 34 & \textbf{378.72}  \\  
 0.6 & 1.1 & 2.9 & 0.5 & 32 & \textbf{499.60}  \\  
 0.6 & 0.6 & 2.3 & 0.6 & 49 & \textbf{657.82} \\   
 \hline 
 \end{tabular}
 \vspace{1em}
 \label{tab:SR_reconstruction}
\end{table*}
\begin{figure*}[t]
\centering
\footnotesize
\begin{tabular}{cc}
  Original & Noisy\\[1mm]
\includegraphics[width=0.45\textwidth] 
  {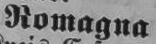} &
\includegraphics[width=0.45\textwidth] 
  {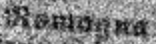}\\
\includegraphics[width=0.45\textwidth] 
  {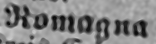} &
\includegraphics[width=0.45\textwidth] 
  {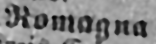}\\[1mm]
  EED, \hspace{-0.5mm} MSE=510.80 &
  SD, \hspace{-0.5mm} MSE=499.60
\end{tabular} 
\vspace{-5pt}  

\caption{\small
Zoom into SR reconstructions for the Text2 dataset with sub-optimal flow.}
\label{fig:text2}
\vspace{1em}
\footnotesize
\begin{tabular}{cccc} 
 \includegraphics[width=0.24\textwidth] {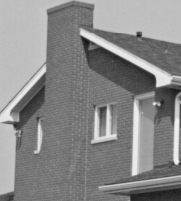} & 
 \includegraphics[width=0.24\textwidth] {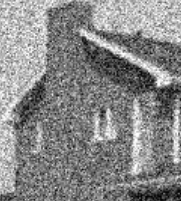} &
 \includegraphics[width=0.24\textwidth] {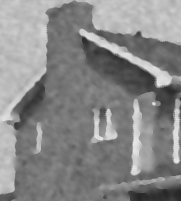} &
 \includegraphics[width=0.24\textwidth] {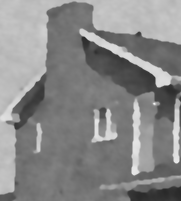}\\[1mm]
 Original & Noisy & EED, \hspace{-0.5mm} MSE=$172.94$ & SD, \hspace{-0.5mm} 
 MSE=$141.62$ 
\end{tabular}
\vspace{-5pt}  
\caption{\small
Zoom into SR reconstructions for the House2 dataset with sub-optimal flow.}
\label{fig:house2}
\end{figure*}
\begin{table}[t]
\caption{\normalsize Data term evaluation. 
\textbf{Left:} Text2 with ground truth flow.
\textbf{Right:} Text2 with sub-optimal flow.}
\vspace{5pt}
\begin{minipage}{0.49\textwidth}
\setlength{\tabcolsep}{3pt}
\small
\centering
\begin{tabular}{lrrrrrr}
\hline
 Model & $\sigma$ & $\;\sigma_B$ & $\;\;\;\lambda$ 
 & $\;\;\;\alpha$ & $k_{\textrm{max}}$ & \;\;\;MSE \\
\hline 
1 & 0.6 & 
1.0 & 2.7 & 0.6 & 34 & \textbf{378.72} \\ [2pt]
2 & 0.6 & 
1.0 & 2.7 & 0.6 & 34 & 382.63 \\ [2pt]
3 & 0.6 & 
0.6 & 2.9 & 1.2 & 20 & 381.75 \\ [2pt]
4 & 0.6 & 
0.8 & 2.8 & 0.6 & 35  & 392.66 \\ [2pt]
5 & 0.6 & 
0.5 & 2.7 & 0.7 & 33 & 391.91 \\ [2pt]
6 & 0.3 & 
0.5 & 4.1 & 0.4 & 60 & 403.50 \\ [2pt]
2.1 & 0.6 & 
1.5 & 3.3 & 0.2 & 55 & 394.85 \\ [2pt]
\hline 
\end{tabular}
\end{minipage}
\hspace{2mm}
\begin{minipage}{0.49\textwidth}
\setlength{\tabcolsep}{3pt}
\small
\centering
\begin{tabular}{lrrrrrr}
\hline 
 Model & $\sigma$ & $\;\sigma_B$ & $\;\;\;\lambda$ 
 & $\;\;\;\alpha$ & $k_{\textrm{max}}$ & \;\;\;MSE \\
\hline 
1 & 0.6 & 
1.1 & 2.9 & 0.5 & 32 & \textbf{499.60} \\ [2pt]
2 & 0.6 & 
1.1 & 3.4 & 0.4 & 33 & 502.78 \\ [2pt]
3 & 0.3 & 
0.8 & 3.8 & 1.0 & 21 & 500.50 \\ [2pt]
4 & 0.6 & 
0.9 & 3.0 & 0.5 & 32  & 511.09 \\ [2pt]
5 & 0.4 & 
0.6 & 4.6 & 0.3 & 43 & 513.29 \\ [2pt]
6 & 0.4 & 
0.6 & 4.6 & 0.3 & 48 & 518.04 \\ [2pt]
2.1 & 0.6 & 
1.6 & 3.5 & 0.2 & 56 & 523.34 \\ [2pt]
\hline
\end{tabular}
\end{minipage}

\label{tab:SR_observational}
\end{table}
\subsection{Denoising Experiments}
\subsubsection{Datasets.} The test images for denoising 
experiments Lena, House, Peppers and 
Bridge\footnote{\url{http://sipi.usc.edu/database/}} 
were corrupted with clipped-AWGN ($\sigma_{\textrm{noise}} =$ 40, 
60 and 80).

\subsubsection{Parameter Selection.} Our experience indicates 
that 36 sectors gives a reasonable directional resolution. 
Also, we have chosen the radius of the disc-shaped 
neighborhood to be 7 pixels. 
The time step size of the explicit scheme for SD was 
chosen such that the maximum--minimum principle is not violated. 
Our grid size is set to $1$. For EED, we choose the time step size
$\tau=0.2$. All other parameters 
(Gaussian smoothing $\sigma$, diffusivity parameter $\lambda$, 
and number of iterations $k_{\textrm{max}}$) 
have been optimised with respect 
to the mean squared error (MSE). 

\subsubsection{Denoising Performance.}
In the first experiment, using equations (\ref{eq:EED}) and 
(\ref{sec_evol}), we evaluate the denoising performance of 
EED and SD, respectively. It is clear from MSE values in 
Table \ref{tab:denoising} 
that SD produces better results. 
The superior performance of SD compared to EED can be attributed to 
its higher adaptivity towards image structures like edges. 
\subsection{Super-resolution Reconstruction Experiments}
\subsubsection{Datasets.} 
For the SR reconstruction experiments,
we have considered two high-resolution scenes in the form of
`Text'\footnote{\url{https://pixabay.com/en/knowledge-book-library-glasses-1052014/}} and `House' images. 
The ground truth HR images have been warped (randomly generated 
deformation motion), 
blurred (Gaussian blur with standard deviation 1.0),
downsampled (with bilinear interpolation), and
degraded by noise (clipped-AWGN with $\sigma_{\textrm{noise}} = $40).  

\subsubsection{Parameter Selection.}
To account for a large spectrum of optical flow qualities, we have
used both the ground truth flow as well as a simplified approach of
Brox et al.~\cite{BBPW04} without gradient constancy assumption.
The parameters for different datasets are shown in Table 
\ref{tab:SR_models}. We optimise these parameters just once, but 
not after every super-resolution iteration.
For SR reconstruction, we additionally optimise 
the parameters $\alpha$ (smoothness) and $\sigma_{\textrm{B}}$ 
(Gaussian blur operator) apart from the already mentioned denoising 
parameters. Again the grid size is $1$. As time step we choose $\tau=0.05$ 
for EED and $\tau=0.012$ for SD, giving experimental stability and
convergence to a plausible reconstruction. 
We initialise $\bm{u}$ with a bilinearly upsampled image.

\subsubsection{Smoothness Term Evaluation.}
The SR reconstruction quality of the 
two regularisers is evaluated using equations (\ref{eq:EED_SR})
and (\ref{eq:SD_SR}). From Table \ref{tab:SR_reconstruction}
and Figures \ref{fig:text2},\ref{fig:house2} 
we observe that SD outperforms EED consistently.
This holds both for ground truth and suboptimal optical flow,
over all downsampling factors.

\subsubsection{Data Term Evaluation.}
Since we have observed a superior performance of SD for 
regularisation purposes, we also use it in the smoothness 
term while evaluating the data term with the observational model.
Table \ref{tab:SR_observational} shows the MSE values of the 
reconstructed high resolution scene with all observational models 
from Table \ref{tab:SR_models}. 

For ground truth flow, the observational model M1 performs best. 
This is in accordance with \cite{WQ2004, BW2017}. 
For suboptimal flow, M1 also outperforms M2. Interestingly, this is
in contrast to the findings in \cite{WQ2004, BW2017}, where M2 gave 
superior results for SR problems without noise. 
We explain this by the fact that we first warp the HR scene in M1.
This introduces an error by applying a motion field computed from 
blurred LR images to sharp HR images. 
On the other hand, such an error does not occur for M2, as we 
first blur the HR scene. However, swapping blur and warp operators 
induces errors since matrix multiplication is not commutative. 
The error magnitude depends on the images and their noise.
In our case, we conjecture that the latter error is higher than the 
former. Therefore, M1 outperforms M2.

In \cite{BW2017}, model M2.1 was much faster than M2 with only
little loss in reconstruction quality. 
However, this model becomes irrelavent in the noisy scenario, as 
M1 outperforms M2, and we also encounter a further quality loss  
when replacing M2 by M2.1.
\section{Conclusion and Outlook}
\label{sec:conclusion}

Our paper belongs to the scarce amount of literature that ventures 
to investigate super-resolution models in the practically relevant
scenario of substantial amounts of clipped noise. In contrast to classical
least squares approaches with homogeneous diffusion regularisation
we have paid specific attention to structure preserving regularisers
such as edge-enhancing anisotropic diffusion (EED). Interestingly,
EED has not been used for super-resolution before, in spite of the fact
that alternatives such as BM3D and NLB are less suited for super-resolution 
from data with clipped noise. More importantly, we have also proposed
a novel anisotropic diffusion model called sector diffusion. It is
the first diffusion method that consequently uses only one-sided 
directional derivatives. In its local formulation, this is a model
that offers also structural novelties from a mathematical perspective,
since it cannot be described in terms of a partial differential equation.
From a practical perspective, the non-local sector diffusion possesses a 
higher structural adaptivity and a better denoising performance than 
simpler diffusion models. 
Thus, it appears promising to study its usefulness also in applications 
beyond super-resolution. This is part of our ongoing work.

\noindent {\bf Acknowledgements.}
J.W. has received funding from the European Research 
Council (ERC) under the European Union's Horizon 2020 research 
and innovation programme (grant no. 741215, ERC Advanced Grant 
INCOVID). We thank our colleagues Dr. Matthias Augustin and 
Dr. Pascal Peter for useful comments on a draft version of 
the paper.

%
%
%
\bibliographystyle{splncs04}
\bibliography{myrefs}

\begin{thebibliography}{10}
\providecommand{\url}[1]{\texttt{#1}}
\providecommand{\urlprefix}{URL }
\providecommand{\doi}[1]{https://doi.org/#1}

\bibitem{BW2017}
Bodduna, K., Weickert, J.: Evaluating data terms for variational multi-frame
  super-resolution. In: Lauze, F., Dong, Y., Dahl, A. (eds.) Scale Space and
  Variational Methods in Computer Vision. Lecture Notes in Computer Science,
  vol. 10302, pp. 590--601. Springer, Berlin (Jun 2017)

\bibitem{BBPW04}
Brox, T., Bruhn, A., Papenberg, N., Weickert, J.: High accuracy optical flow
  estimation based on a theory for warping. In: Pajdla, T., Matas, J. (eds.)
  Computer Vision -- {ECCV} 2004, Part IV, Lecture Notes in Computer Science,
  vol.~3024, pp. 25--36. Springer, Berlin (2004)

\bibitem{DFKE07}
Dabov, K., Foi, A., Katkovnik, V., Egiazarian, K.: Image denoising by sparse
  {3D} transform-domain collaborative filtering. IEEE Transactions on Image
  Processing  \textbf{16}(8),  2080--2095 (Aug 2007)

\bibitem{DDICGDPSI2015}
Doulamis, A., Doulamis, N., Ioannidis, C., Chrysouli, C., Grammalidis, N.,
  Dimitropoulos, K., Potsiou, C., Stathopoulou, E., Ioannides, M.: 5d
  modelling: {A}n efficient approach for creating spatiotemporal predictive 3d
  maps of large-scale cultural resources. ISPRS Annals of the Photogrammetry,
  Remote Sensing and Spatial Information Sciences  \textbf{2}(5),  61--68 (Jan
  2015)

\bibitem{EF1997}
Elad, M., Feuer, A.: Restoration of a single superresolution image from several
  blurred, noisy, and undersampled measured images. IEEE Transactions on Image
  Processing  \textbf{6}(12),  1646--1658 (Dec 1997)

\bibitem{FREM2004a}
Farsiu, S., Robinson, M.D., Elad, M., Milanfar, P.: Fast and robust multi-frame
  super resolution. IEEE Transactions on Image Processing  \textbf{13}(10),
  1327--1364 (Oct 2004)

\bibitem{GSI09}
Glasner, D., Shai, B., Irani, M.: Super-resolution from a single image. In:
  Proc.~IEEE International Conference on Computer Vision (ICCV). pp. 349--356.
  Kyoto, Japan (Sep 2009)

\bibitem{KBPS2011}
Knoll, F., Bredies, K., Pock, T., Stollberger, R.: Second-order total
  generalized variation ({TGV}) for {MRI}. Magnetic Resonance in Medicine
  \textbf{65}(2),  480--491 (Feb 2011)

\bibitem{KDV2012}
Kosmopoulos, D., Doulamis, N., Voulodimos, A.: Bayesian filter based behavior
  recognition in workflows allowing for user feedback. Computer Vision and
  Image Understanding  \textbf{116},  422--434 (Mar 2012)

\bibitem{LHR2015}
Laghrib, A., Hakim, A., Raghay, S.: A combined total variation and bilateral
  filter approach for image robust super resolution. EURASIP Journal on Image
  and Video Processing  \textbf{2015}(1), ~19 (June 2015)

\bibitem{LBM2013}
Lebrun, M., Buades, A., Morel, J.: A nonlocal {B}ayesian image denoising
  algorithm. SIAM Journal on Imaging Sciences  \textbf{6}(3),  1665--1688 (Sep
  2013)

\bibitem{LMZBBGAC2013}
Li, X., Mooney, P., Zheng, S., Booth, C., Braunfeld, M., Gubbens, S., Agard,
  D., Cheng, Y.: Electron counting and beam-induced motion correction enable
  near-atomic-resolution single-particle cryo-{EM}. Nature Methods
  \textbf{10}(6),  584--590 (Jun 2013)

\bibitem{LFCS2005}
Lin, F., Fookes, C., Chandran, V., Sridharan, S.: Investigation into optical
  flow super-resolution for surveillance applications. In: Lovell, B.C.,
  Maeder, A.J. (eds.) APRS Workshop on Digital Image Computing: Pattern
  Recognition and Imaging for Medical Applications. pp. 73--78. Brisbane (Feb
  2005)

\bibitem{MKDD2015}
Makantasis, K., Karantzalos, K., Doulamis, A., Doulamis, N.: Deep supervised
  learning for hyperspectral data classification through convolutional neural
  networks. In: Proc.~IEEE International Geoscience and Remote Sensing
  Symposium (IGARSS). pp. 4959--4962 (Jul 2015)

\bibitem{MO2008}
Marquina, A., Osher, S.: Image super-resolution by {TV}-regularization and
  {B}regman iteration. Journal of Scientific Comuputing  \textbf{37}(3),
  367--382 (Dec 2008)

\bibitem{PVS2006}
Pham, T., Vliet, L., Schutte, K.: Robust fusion of irregularly sampled data
  using adaptive normalized convolution. EURASIP Journal on Advances in Signal
  Processing  \textbf{2006}(083268) (Dec 2006)

\bibitem{TLAN2001}
Tatem, A., Lewis, H., Atkinson, P., Nixon, M.: Super-resolution target
  identification from remotely sensed images using a {H}opfield neural network.
  IEEE Transactions on Geoscience and Remote Sensing  \textbf{39}(4),  781--796
  (Apr 2001)

\bibitem{TLAN2002}
Tatem, A., Lewis, H., Atkinson, P., Nixon, M.: Super-resolution land cover
  pattern prediction using a {H}opfield neural network. Remote Sensing of
  Environment  \textbf{79}(1),  1--14 (Jan 2002)

\bibitem{WQ2004}
Wang, Z., Qi, F.: On ambiguities in super-resolution modeling. IEEE Signal
  Processing Letters  \textbf{11}(8),  678--681 (Aug 2004)

\bibitem{We94a}
Weickert, J.: Anisotropic diffusion filters for image processing based quality
  control. In: Fasano, A., Primicerio, M. (eds.) Proc.~Seventh European
  Conference on Mathematics in Industry, pp. 355--362. Teubner, Stuttgart
  (1994)

\bibitem{We94e}
Weickert, J.: Theoretical foundations of anisotropic diffusion in image
  processing. In: Kropatsch, W., Klette, R., Solina, F., Albrecht, R. (eds.)
  Theoretical Foundations of Computer Vision, Computing Supplement, vol.~11,
  pp. 221--236. Springer, Vienna (1996)

\bibitem{We97}
Weickert, J.: Anisotropic Diffusion in Image Processing. Teubner, Stuttgart
  (1998)

\bibitem{WWW2013}
Weickert, J., Welk, M., Wickert, M.: L2-stable nonstandard finite differences
  for anisotropic diffusion. In: Kuijper, A., Bredies, K., Pock, T., Bischof,
  H. (eds.) Scale Space and Variational Methods in Computer Vision. pp.
  380--391. Lecture Notes in Computer Science, Springer, Berlin (Jun 2013)

\bibitem{YWHM10}
Yang, J., Wright, J., Huang, T., Ma, Y.: Image super-resolution via sparse
  representation. IEEE Transactions on Image Processing  \textbf{19}(11),
  2681--2873 (May 2010)

\bibitem{YZS2012}
Yuan, Q., Zhang, L., Shen, H.: Multiframe super-resolution employing a
  spatially weighted total variation model. IEEE Transactions on Circuits,
  Systems and Video Technology  \textbf{22}(3),  379--392 (Mar 2012)

\end{thebibliography}

\end{document}